\documentclass{article}
\usepackage{ijcai22}

\usepackage{times}
\usepackage{soul}
\usepackage{url}
\usepackage[hidelinks]{hyperref}
\usepackage[utf8]{inputenc}
\usepackage[small]{caption}
\usepackage{graphicx}
\usepackage{amsmath}
\usepackage{amsthm}
\usepackage{booktabs}
\usepackage{algorithm}
\usepackage{algorithmic}

\usepackage{amsthm,amsmath,amssymb}

\title{DMS-GCN: Dynamic Mutiscale Spatiotemporal Graph Convolutional Networks for Human Motion Prediction}

\author{
Zigeng Yan\and
Di-Hua Zhai\footnote{Corresponding author.}\and
Yuanqing Xia\\
\affiliations
School of Automation, Beijing Institute of Technology\\
\emails
yanzigeng@bit.edu.cn,~zhaidih@bit.edu.cn,~xia$\_$yuanqing@bit.edu.cn
}

\begin{document}

\maketitle

\begin{abstract}
  Human motion prediction is an important and challenging task in many computer vision application domains. Recent work concentrates on utilizing the timing processing ability of recurrent neural networks (RNNs) to achieve smooth and reliable results in short-term prediction. However, as evidenced by previous work, RNNs suffer from errors accumulation, leading to unreliable results. In this paper, we propose a simple feed-forward deep neural network for motion prediction, which takes into account temporal smoothness and spatial dependencies between human body joints. We design a Multi-scale Spatio-temporal graph convolutional networks (GCNs) to implicitly establish the Spatio-temporal dependence in the process of human movement, where different scales fused dynamically during training. The entire model is suitable for all actions and follows a framework of encoder-decoder. The encoder consists of temporal GCNs to capture motion features between frames and semi-autonomous learned spatial GCNs to extract spatial structure among joint trajectories. The decoder uses temporal convolution networks (TCNs) to maintain its extensive ability. Extensive experiments show that our approach outperforms SOTA methods on the datasets of Human3.6M and CMU Mocap while only requiring much lesser parameters. Code will be available at \url{https://github.com/yzg9353/DMSGCN}.
\end{abstract}

\section{Introduction}

It is an innate instinct for humans to predict the laws of the movements of things, especially human motion. The fantastic predictive ability helps one avoid danger or win in sports competitions. The use of computers to perceive human behavior is still in its infancy, how to predict human behavior has become a hot topic in that field.

\begin{figure}[!t]
	\centering
	\includegraphics[width=3.2in]{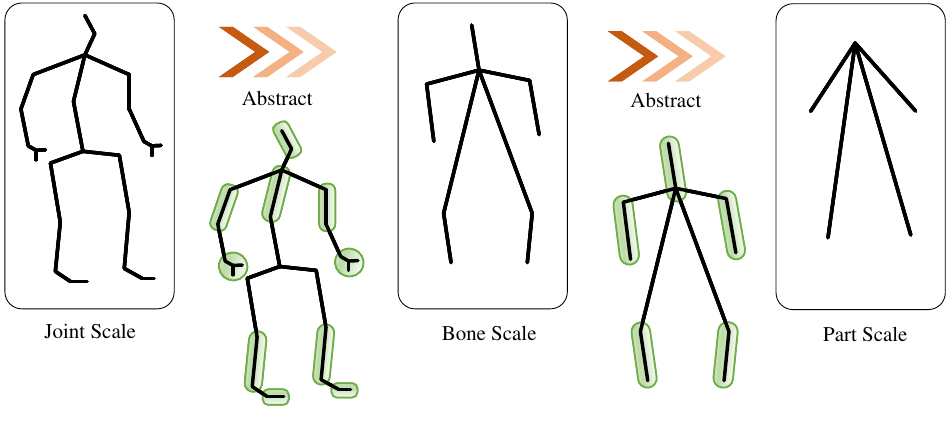}
	\caption{\textbf{Human pose divided into three scales.} We abstract the human pose from fine to coarse into three scales: the Joint Scale, the Bone Scale, and the Part Scale. As for the downsampling method, we determine the clustering form according to subjective experience and use the mean value of the selected joints to generate a new part.}
	\label{scale}
\end{figure}

The spatial characteristics of human motion are mainly reflected in the interdependence of human body joints. Therefore, how to extract the spatial dependencies between different joints becomes crucial. The network model based on the human kinematic chain \cite{aksan2019structured} and the direct use of graph convolution networks \cite{li2018convolutional,mao2019learning,cui2020learning} become the two main directions of spatial dependence extraction. Feature extraction along the human kinematic chain tends to suffer message decay for two joints with long distances among them. Graph convolution networks often set adjacency matrix according to human body structure, in which a joint usually receives information among its neighbor joints with the same weight. However, as for human motion, the connection degrees between joints are unmanageable to determine artificially. Therefore, we propose a new method that creates a semi-autonomous self-learning graph convolution limited by a mask. Our method ensures that joint information does not propagate along the human kinematic chain to avoid information loss, the information transmission channel of the distant joints is closed to prevent the network from converging to the local optimum, and the networks automatically learn the connection strength between joints in a neighborhood.

Because of the temporal nature of human motion, the most common trend is using recurrent neural networks (RNNs) \cite{fragkiadaki2015recurrent,jain2016structural,martinez2017human,pavllo2020modeling} or Transformers\cite{cai2020learning}. As demonstrated by previous work \cite{mao2019learning,li2018convolutional}, RNNs suffer from two main shortcomings. First, the existing work uses the estimation of the current step of RNNs as the input of the next step, so the network tends to accumulate errors in the generated sequence, which eventually leads to the collapse of the result. Secondly, RNNs tend to lose information in transmission. Temporal convolution networks (TCNs) recently achieved better results than RNNs on many tasks due to their parallel training with a flexible receptive field. In this paper, we use TCNs as our decoder to generate the future frame due to their performance and robustness. As for the encoder, we define the temporal edges, which connect joints across consecutive time steps, to utilize temporal graph convolution networks so that information can be integrated along the temporal dimension.

Human joints move in small groups called body parts when performing activities. Skeleton-based action recognition approaches \cite{yan2018spatial} and motion prediction approaches \cite{li2020dynamic,dang2021msr} have verified the effectiveness of introducing body parts into modeling human motion. So we abstract the human pose from fine to coarse into three levels: joint scale, bone scale, and part scale, in which the fusion procedures between different scales are learned by networks dynamically. In this way, the part scale restricts the modeling of the bone scale trajectories within local regions, then the bone scale can further constrain the joint scale trajectories, thus forming a robust method to extract the hierarchical representation from the skeleton sequences.

The principal contributions of this paper are summarized as:

(i) We abstract the human pose into joint scale, bone scale, and part scale, and fuse them dynamically, which enhances the extracting ability of the encoder to ensure the predicted human pose is stable.

(ii) We combine graph convolution networks with the human kinematic chain to generate a revised graph convolution network with semi-autonomous-learned adjacency, which is proved to be helpful by our ablation experiments.

(iii) We define joint connection across consecutive time steps as the temporal edges to utilize temporal graph convolution networks, allowing temporal information flow across frames.

(iv) Experiments on standard human motion prediction benchmarks demonstrate the superiority of our approach; our model outperforms SOTA methods on the datasets of Human3.6M and CMU Mocap while only requiring much lesser parameters.

\section{Related Works}
\textbf{Human motion prediction at early stage.} Human motion prediction has been an important issue in the research field of computer vision since the beginning of this century and mainly used by data driven methods, such as conditional Boltzmann machine implicit hybrid model \cite{taylor2010dynamical}, bilinear spatio-temporal basis model \cite{akhter2012bilinear} and nonparametric sparse Bayesian network model \cite{lehrmann2013non}. The above methods using time series processing means to effectively predict human periodic motion and simple aperiodic motion, but they do not perform well for some more complex and changeable human actions. In recent years, the emergence of large-scale human motion pose datasets and deep learning methods gives hope to gradually solve those problems.

\textbf{RNN-based human motion prediction models.} Because of the success at natural language processing field \cite{chelba2013one,chung2014empirical}, and also the temporal nature of human motion data, the most common trend is using recurrent neural networks (RNNs) \cite{fragkiadaki2015recurrent,jain2016structural,martinez2017human,pavllo2020modeling}. ERD \cite{fragkiadaki2015recurrent} introduces RNN structure into human motion prediction, which verifies the effectiveness of RNN. Error accumulation is inevitable, and a curriculum learning strategy is used during the training process to reduce its influence. Res-sup \cite{martinez2017human} is a simple yet effective model using Seq2Seq and the discontinuity of the initial prediction frame due to the residual velocities architecture in its decoder part. To further improve the accuracy, a large number of improvements are tried based on Res-sup. QuatNet \cite{pavllo2020modeling} converts the input data format from rotation vector to quaternion to avoid the cumulative error of rotation vector along the kinematic chain. ConvSeq2Seq \cite{li2018convolutional} introduces a convolution layer into the recurrent neural network to extract spatial information about human bone structure. HMR \cite{liu2019towards} extends the GRU block into two-dimension and extracted spatiotemporal dependence at the same time. PVRED \cite{wang2021pvred} concatenates joint velocities, frame position, and joint position to enrich network input information. SPL \cite{aksan2019structured} designs a structured prediction layer for motion extraction. These approaches above are of great significance for the follow-up work, but the error accumulation characteristic of RNNs makes them unsuitable for long-term prediction.

\textbf{Extraction of human spatial information.} Human motion data contains both temporal and spatial dimensions, so it is crucial to extract information along both dimensions. RNN-based models (as listed above) usually lack the extraction of human spatial information. In recent years, extracting human spatial information has become a research hotspot. SPL \cite{aksan2019structured} mentioned before encodes human motion by exploiting the kinematic tree of the human body. The use of a kinematic tree does not perfectly reflect the real dependency, as discussed in \cite{li2018convolutional}, each movement requires synchronizing different body parts, even distant ones indirectly connected by the kinematic tree. The use of graph convolution can realize the information transmission of indirectly connected joints by increasing the max-hop of information transmission. LearnTrajDep \cite{mao2019learning} introduced GCN into human motion prediction earlier and built a feedforward network to predict the changes of joints over time in the trajectory space. DMGNN \cite{li2020dynamic} convolved the human body joints in three scales and dynamically adjusted the connection weight of each layer. STSGCN \cite{sofianos2021space} tries to solve the overlapping issue for space-time information by applying space-time-separable graph convolutional networks. MSR-GCN \cite{dang2021msr} utilizes graph convolution networks to extract spatial information from fine to coarse body scale. LearnTrajDep needs many GCN layers, and MSR-GCN has many handmade structures, and they both consume vast quantities of computer resources. In our work, we will propose a weight-light but acts-best method.

\begin{figure*}[!t]
	\centering
	\includegraphics[width=6.5in]{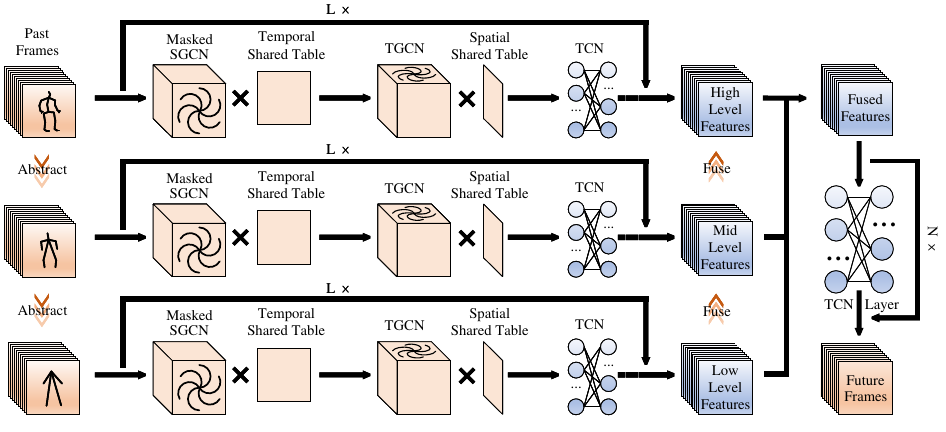}
	\caption{\textbf{Overall architecture of our method.} We apply spatial graph convolution and temporal graph convolution simultaneously for each scale to extract spatiotemporal information. Then the resulting hidden features in different levels fuse dynamically by a Linear network. Finally, a series of TCN layers perform as a generator to predict future frames.}
	\label{architecture}
\end{figure*}

\section{DMS-GCN}
\subsection{Problem Formulation}
By observing the body pose given by the 3D coordinates of $ V $ human joints for $ T $ frames, the model needs to predict the pose for $ K $ future frames.

Supposed that an observed pose sample can be formulated as
$ \textbf{X}_{1:T}=\{\textbf{x}_{1},\textbf{x}_{2},\dots,\textbf{x}_{t},\dots,\textbf{x}_{T}\} $ with $ 1 \leqslant t \leqslant T $ where a frame $ \textbf{x}_{t} \in \mathbb{R}^{N} $ at time $ t $ denotes the $ N $-dimensional body pose. $ N = V \times C $, where $ N $ is the number of joints in the skeleton representation unique in different scales (we use 22, 10, and 5 joints for our three scales in this paper) and $ C $ the dimension of each joint. Following the previous works, each joint of the 3D skeleton is represented as 3D coordinates, thus we have $ C=3 $. Then, we denote $ {\hat{\textbf{Y}}}_{T+1:T+K} = \{ {\hat{\textbf{y}}}_{T+1}, {\hat{\textbf{y}}}_{T+2}, \dots, {\hat{\textbf{y}}}_{T+K} \} $ be the prediction frames, and $ {\textbf{Y}}_{T+1:T+K} = \{ {\textbf{y}}_{T+1}, {\textbf{y}}_{T+2}, \dots, {\textbf{y}}_{T+K} \} $ be the corresponding ground truth.

For the convenience of calculation, the input motion history data $ \mathbf{I} $ was formulated as a $ N \times C \times T \times V $ shaped tensor. For spatial graph convolution, we define a graph $ \mathcal{G}_s = (\mathcal{V}_s, \mathcal{E}_s) $ with $ V $ nodes $ i \in \mathcal{V}_s $. Edges $ (i, j) \in \mathcal{E}_s $ are represented by a masked spatial adjacency $ \mathbf{A}_{s} \in \mathbb{R}^{V \times V} $. Similarly for temporal graph convolution, we define another graph $ \mathcal{G}_t = (\mathcal{V}_t, \mathcal{E}_t) $ with $ T $ nodes $ t \in \mathcal{V}_t $. Temporal edges $ (t, u) \in \mathcal{E}_t $ are represented by a temporal adjacency $ \mathbf{A}_{t} \in \mathbb{R}^{V \times V}$.

\subsection{Dynamic Mutiscale Human Pose}

The human pose can be abstract progressively to obtain a set of scaled poses. Intuitively, two ideas motivate us to propose a multiscale architecture to leverage the information from different scales. First, it is much easier to make predictions on a coarse scale than on the fine one. What's more, the coarse one restricts the modeling of the fine-scale trajectories within local regions, which facilitates the stableness of predicted poses.

As shown in Figure \ref{scale}, the original scale called ``Joint Scale'' has twenty-two joints. We abstract the original scale recursively to obtain two scales, the ``Bone Scale'' with ten joints and the ``Part Scale'' with five joints. The downsampling methods drawn between scales depict how to combine the joints in the fine-level scale. We take an average 3D coordinate as the place of the new joints. Note that the grouping method we use is the most stable one obtained after several trials.

\textbf{Dynamic fusing process.} After information extraction by GCNs, we obtain features at three levels. It is crucial to fuse the lower level features to a higher level to facilitate the abundance of information for the decoder input. We tried some manually designed ways that do not perform well, mainly because their rigid structures hinder the convergence of the model where the parameters are updated simultaneously. Then we discover that only a Linear layer seems to work well. According to the results, it is plausible to fine tune the fusing way dynamically by networks. The fusing process can be expressed by the following formula:
\begin{align}
	\mathbf{X}_{2}^{+} &= \alpha \mathbf{W}_{32}\mathbf{X}_3 + (1 - \alpha) \mathbf{X}_2, \\
	\mathbf{X}^{+} &= \alpha \mathbf{W}_{21}\mathbf{X}_{2}^{+} + (1 - \alpha) \mathbf{X}_1,
\end{align}
where $ X_i $ represents the features in the $ i $th level in Figure \ref{architecture} from top to bottom, $ W_{ij} $ the weight matrix upsampling features from $ i $th level to $ j $th level, $ \alpha $ the fusion coefficient. The superscript ``+'' stands for fused features.

\subsection{Semi-autonomous-learned Spatial GCN}

We use GCNs to extract the spatial information of different scales. Intuitively, relying on a predefined adjacent matrix \cite{yan2018spatial} according to the human kinematic chains neglects the intensity of information interaction between joints. The method \cite{mao2019learning} regards the adjacency matrix as a parameter to train the connection strength has too many redundant connections, which lead the network hard to convergence. To overcome those drawbacks, we propose a semi-autonomous-learned GCN network.

\textbf{Background on GCNs.} As described above, we assume that the human body pose as a graph $ \mathcal{G}_s = (\mathcal{V}_s, \mathcal{E}_s) $ with $ V $ nodes $ i \in \mathcal{V}_s $. Edges $ (i, j) \in \mathcal{E}_s $ are represented by an adjacency $ \mathbf{A}_{s} \in \mathbb{R}^{V \times V} $. Then, a pose sequence with $ T $ frames is formulated as $ \mathcal{M}_s=\{ \mathcal{G}_1, \mathcal{G}_2, \dots, \mathcal{G}_T \} $. Take one frame as an example, a graph convolutional layer $ p $ then takes as input a matrix $ \mathbf{H}_{s}^{(p)} \in \mathbb{R}^{V \times C^{(p)}} $ with $ C^{(p)} $ the graph node dimensions of layer $ p $. Given this information and a set of trainable weights $ \mathbf{W}_{s}^{(p)} \in \mathbb{R}^{C^{(p)} \times C^{(p+1)}} $, we describe a graph convolutional layer as the form
\begin{align}
	\mathbf{H}_{s}^{(p+1)} = \sigma ( \mathbf{A}_{s}^{(p)} \mathbf{H}_{s}^{(p)} \mathbf{W}_{s}^{(p)} ),
\end{align}
where $ \sigma $ is an activation function, such as ReLU or PReLU.

\begin{figure}[!t]
	\centering
	\includegraphics[width=3.2in]{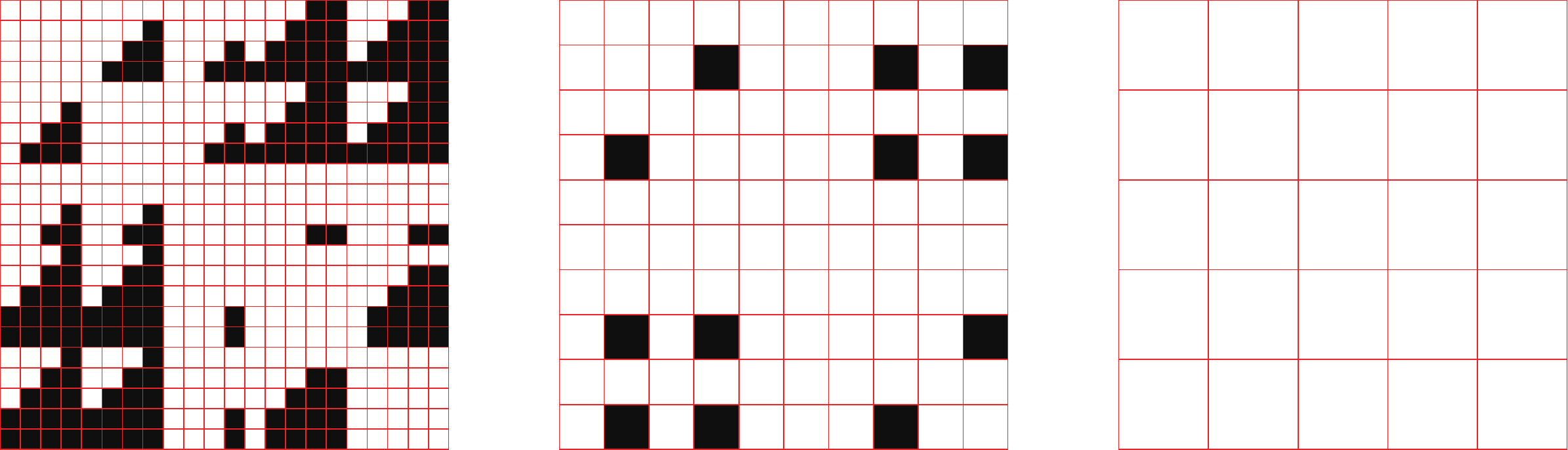}
	\caption{\textbf{Masks used in different scales.} From left to right, this figure shows the masks used in the SGCN process for ``Joint Scale'', ``Bone Scale'' and ``Part Scale'' respectively. The white block with the value ``1'' means connected, and the black block with the value ``0'' means disconnected.}
	\label{A}
\end{figure}

\textbf{Our improvements.} Instead of relying on a fully connective adjacency as in \cite{mao2019learning}, we decide to restrict the joint within several hops. So we design a fixed mask matrix $ \mathbf{M} \in \mathbb{R}^{V\times V} $ to block the connection of some joint pairs with long distances. Besides, we also design a Time-Sharing Weight Table $ \mathbf{T}_{s}^{(p)} $ which contains the learned spatio feature to embed the joints. Then the general formula of a GCN layer becomes
\begin{align}
 	\mathbf{H}_{s}^{(p+1)} = \sigma ( \mathbf{T}_{s}^{(p)}(\mathbf{A}_{s}^{(s)} \circ \mathbf{M}) \mathbf{H}_{s}^{(p)} \mathbf{W}_{s}^{(p)} ),
\end{align}
where $ \mathbf{M} $ changes according to the pose graph, i.e., each scale we use has its mask, as shown in Figure \ref{A}. The symbol $ \circ $ indicates an element-wise product. Masks prevent the connection of some joints and reduce the updatable parameters of the adjacency matrix, making the adjacency matrix semi-autonomous learned.

\subsection{Temporal GCN}

Similar to the basic Spatio GCN, we define temporal edges that connect joints across the frame. Then we obtain a temporal graph $ \mathcal{G}_t = (\mathcal{V}_t, \mathcal{E}_t) $ with $ T $ nodes $ t \in \mathcal{V}_t $. Temporal edges $ (t, u) \in \mathcal{E}_t $ are represented by a temporal adjacency $ \mathbf{A}_{t} \in \mathbb{R}^{V \times V} $. This time, we only add a Space-Sharing Weight Table $ \mathbf{T}_{t}^{(p)} $ contains the learned temporal features:
\begin{align}
	\mathbf{H}_{t}^{(p+1)} = \sigma ( \mathbf{T}_{t}^{(p)} \mathbf{A}_{t}^{(p)} \mathbf{H}_{t}^{(p)} \mathbf{W}_{t}^{(p)} ),
\end{align}
where $ \mathbf{H}_{t}^{(p)} \in \mathbb{R}^{T\times C^{(p)}} $ is the input matrix of layer $ p $, and $ \mathbf{W}_{t}^{(p)} \in \mathbb{R}^{C^{(p)}\times C^{(p+1)}} $ is the trainable weight matrix of layer $ p $.

\subsection{Decoding Future Poses}
Once we get the body dynamics features from the encoder, the estimation of 3D coordinates of body joints in the future falls to the decoder that applies to the temporal dimension. The decoder takes the observed frames as the input and the body dynamics features as the rule to generate plausible future poses.

To tackle the training difficulty and mitigate the error accumulation of traditional RNN \cite{fragkiadaki2015recurrent,jain2016structural,martinez2017human,pavllo2020modeling}, we adopt Temporal Convolution Networks (TCN) \cite{sofianos2021space} as our decoder.

\section{Experiments}
\subsection{Datasets and Experimental Setup}

\textbf{Human 3.6M.} Following previous work, we use the Human3.6M dataset (H3.6M) \cite{ionescu2013human3}, which is widely used in training for human motion prediction models. H3.6M includes 15 identical actions such as walking, eating, sitting, and phoning, which are performed by 7 actors. For a fair comparison, we adopted the representation with 3D coordinates from \cite{mao2019learning,dang2021msr}, where each pose contains 32 joints. We report the same average error as \cite{mao2019learning,dang2021msr} do on all test sequences.

\textbf{CMU motion capture.} To prove the universality of our model, we also report results on the CMU motion capture dataset (CMU Mocap). CMU Mocap has captured different actions with more joints in comparison with H3.6M. We apply the same preprocessing procedure as on H3.6M. For a fair comparison, we adopted the same data representation, training/test splits, and evaluation sequences as in \cite{mao2019learning}.

\textbf{Model configuration.} We implement our model with Pytorch 1.9.0 on one NVIDIA Quadro RTX 6000 GPU. In the encoder, we use 7 STGCN Blocks, each of them consisting of a proposed Masked SGCN layer and a TGCN layer, with a dropout rate of 0.1. Besides, we use residual layers between STGCN blocks to transfer information between them, as shown in Figure \ref{architecture}. Then, we use 4 TCN layers to generate the future frames. In training, we set batch size 32; we use an ADAM optimizer and a scheduler to half the learning rate about every ten epochs.

\begin{table*}[!t]
	\scriptsize
	\centering
	\caption{Comparisons for short-term prediction on 15 action categories of H3.6M and the averages. The best results are highlighted in bold.}
	\renewcommand{\arraystretch}{1}
	\setlength\tabcolsep{3pt}
	\resizebox{17.5cm}{4cm}
	{\begin{tabular}{c|cccc|cccc|cccc|cccc}
			\hline
			Action & \multicolumn{4}{c|}{Walking}  & \multicolumn{4}{c|}{Eating}   & \multicolumn{4}{c|}{Smoking}  & \multicolumn{4}{c}{Discussion} \\
			\hline
			Millisecond (ms) & 80    & 160   & 320   & 400   & 80    & 160   & 320   & 400   & 80    & 160   & 320   & 400   & 80    & 160   & 320   & 400 \\
			\hline
			Residual sup. & 21.82  & 40.02  & 64.49  & 69.47  & 17.24  & 33.72  & 61.19  & 72.54  & 22.32  & 42.32  & 73.71  & 85.27  & 26.45  & 49.75  & 78.44  & 85.04  \\
			Traj-GCN & 11.88  & 22.17  & 38.38  & 44.46  & 8.18  & 16.72  & 32.82  & \textbf{ 40.06 } & 7.83  & 16.01  & 31.54  & 40.06  & 12.23  & 27.06  & 58.90  & 72.68  \\
			MSR-GCN & 14.09  & 24.97  & 43.65  & 50.80  & 9.78  & 19.28  & 36.81  & 44.40  & 9.39  & 18.10  & 34.20  & 41.21  & 13.52  & 29.08  & 60.40  & 73.26  \\
			DMS-GCN & \textbf{ 10.33 } & \textbf{ 20.26 } & \textbf{ 36.40 } & \textbf{ 43.77 } & \textbf{ 7.21 } & \textbf{ 15.98 } & \textbf{ 32.49 } & 40.45  & \textbf{ 6.93 } & \textbf{ 15.06 } & \textbf{ 30.63 } & \textbf{ 37.87 } & \textbf{ 10.09 } & \textbf{ 24.80 } & \textbf{ 56.67 } & \textbf{ 70.77 } \\
			\hline
			\hline
			Action & \multicolumn{4}{c|}{Directions} & \multicolumn{4}{c|}{Greeting} & \multicolumn{4}{c|}{Phoning}  & \multicolumn{4}{c}{Posing} \\
			\hline
			Millisecond (ms) & 80    & 160   & 320   & 400   & 80    & 160   & 320   & 400   & 80    & 160   & 320   & 400   & 80    & 160   & 320   & 400 \\
			\hline
			Residual sup. & 31.53  & 52.67  & 74.32  & 89.23  & 32.82  & 61.98  & 115.32  & 132.76  & 28.36  & 55.59  & 103.13  & 117.32  & 37.72  & 71.35  & 130.55  & 151.35  \\
			Traj-GCN & 8.79  & 19.63  & 43.63  & 54.48  & 17.81  & 37.24  & 76.18  & 92.35  & 10.13  & 20.83  & 42.20  & 51.89  & 13.18  & 29.13  & 66.17  & \textbf{ 83.96 } \\
			MSR-GCN & 9.90  & 21.52  & 45.70  & 56.05  & 18.65  & 40.77  & 83.60  & 99.30  & 11.71  & 23.15  & 45.63  & 55.80  & 14.50  & 31.24  & 70.02  & 88.93  \\
			DMS-GCN & \textbf{ 6.96 } & \textbf{ 17.31 } & \textbf{ 40.60 } & \textbf{ 51.29 } & \textbf{ 13.86 } & \textbf{ 33.07 } & \textbf{ 72.67 } & \textbf{ 90.01 } & \textbf{ 8.57 } & \textbf{ 19.06 } & \textbf{ 40.11 } & \textbf{ 50.16 } & \textbf{ 10.72 } & \textbf{ 27.11 } & \textbf{ 65.54 } & 84.07  \\
			\hline
			\hline
			Action & \multicolumn{4}{c|}{Purchases} & \multicolumn{4}{c|}{Sitting}  & \multicolumn{4}{c|}{Sitting Down} & \multicolumn{4}{c}{Taking Photo} \\
			\hline
			Millisecond (ms) & 80    & 160   & 320   & 400   & 80    & 160   & 320   & 400   & 80    & 160   & 320   & 400   & 80    & 160   & 320   & 400 \\
			\hline
			Residual sup. & 34.37  & 61.93  & 90.38  & 101.59  & 36.41  & 71.18  & 123.40  & 139.31  & 37.41  & 69.35  & 125.15  & 148.32  & 22.17  & 41.07  & 72.88  & 85.10  \\
			Traj-GCN & 15.18  & 32.16  & \textbf{ 65.37 } & \textbf{ 79.21 } & 10.36  & 21.35  & \textbf{ 45.52 } & \textbf{ 57.23 } & 15.89  & 30.55  & 60.55  & 74.73  & 9.80  & 20.70  & \textbf{ 44.72 } & 56.61  \\
			MSR-GCN & 16.50  & 35.19  & 70.98  & 85.66  & 12.08  & 24.36  & 49.56  & 61.20  & 18.77  & 35.76  & 68.28  & 82.63  & 11.26  & 22.96  & 47.80  & 59.57  \\
			DMS-GCN & \textbf{ 12.97 } & \textbf{ 30.89 } & 66.53  & 81.19  & \textbf{ 9.55 } & \textbf{ 20.73 } & 45.61  & 57.31  & \textbf{ 14.75 } & \textbf{ 29.23 } & \textbf{ 59.68 } & \textbf{ 74.07 } & \textbf{ 8.88 } & \textbf{ 20.06 } & 44.87  & \textbf{ 56.42 } \\
			\hline
			\hline
			Action & \multicolumn{4}{c|}{ Waiting} & \multicolumn{4}{c|}{Walking Dog} & \multicolumn{4}{c|}{Walking Together} & \multicolumn{4}{c}{Average} \\
			\hline
			Millisecond (ms) & 80    & 160   & 320   & 400   & 80    & 160   & 320   & 400   & 80    & 160   & 320   & 400   & 80    & 160   & 320   & 400 \\
			\hline
			Residual sup. & 24.61  & 50.95  & 107.58  & 128.48  & 55.12  & 89.24  & 131.99  & 150.08  & 24.65  & 46.54  & 73.58  & 86.51  & 30.20  & 55.84  & 95.07  & 109.49  \\
			Traj-GCN & 11.01  & 23.20  & 48.93  & 60.31  & 22.68  & 44.93  & 82.56  & \textbf{ 95.75 } & 10.30  & 21.01  & 38.73  & 45.60  & 12.35  & 25.51  & 51.75  & 63.29  \\
			MSR-GCN & 12.30  & 25.27  & 51.77  & 63.21  & 23.34  & 48.64  & 91.57  & 106.20  & 11.83  & 22.55  & 41.43  & 48.63  & 13.84  & 28.19  & 56.09  & 67.79  \\
			DMS-GCN & \textbf{ 8.95 } & \textbf{ 20.43 } & \textbf{ 44.61 } & \textbf{ 56.15 } & \textbf{ 19.11 } & \textbf{ 41.49 } & \textbf{ 82.07 } & 99.04  & \textbf{ 8.75 } & \textbf{ 18.67 } & \textbf{ 35.28 } & \textbf{ 42.73 } & \textbf{ 10.51 } & \textbf{ 23.61 } & \textbf{ 50.25 } & \textbf{ 62.35 } \\
			\hline
	\end{tabular}}
	\label{h36m_short}
\end{table*}
\begin{table*}[htbp]
	\scriptsize
	\centering
	\caption{Comparisons for long-term prediction on 5 action categories of H3.6M and the averages. The best results are highlighted in bold.}
	\renewcommand{\arraystretch}{1}
	\setlength\tabcolsep{5pt}
	\resizebox{17.5cm}{1.1cm}
	{\begin{tabular}{c|cc|cc|cc|cc|cc|cc}
		\hline
		Action & \multicolumn{2}{c|}{Walking} & \multicolumn{2}{c|}{Directions} & \multicolumn{2}{c|}{Phoning} & \multicolumn{2}{c|}{Sitting} & \multicolumn{2}{c|}{Takingphoto} & \multicolumn{2}{c}{Average} \\
		\hline
		Millisecond (ms) & 560   & 1000  & 560   & 1000  & 560   & 1000  & 560   & 1000  & 560   & 1000  & 560   & 1000 \\
		\hline
		Residual sup. & 72.70  & 94.27  & 112.76  & 146.42  & 131.99  & 133.57  & 159.51  & 190.93  & 104.67  & 131.71  & 116.33  & 139.38  \\
		Traj-GCN & \textbf{ 53.20 } & \textbf{ 59.88 } & 72.76  & 106.50  & 69.70  & 105.70  & 79.35  & 122.71  & 78.80  & 124.97  & 70.76  & 103.95  \\
		MSR-GCN & 59.35  & 68.12  & 72.80  & 101.94  & 73.08  & 106.78  & 81.03  & 120.67  & 79.69  & 120.39  & 73.19  & 103.58  \\
		DMS-GCN & 55.90  & 69.02  & \textbf{ 69.42 } & \textbf{ 101.55 } & \textbf{ 68.03 } & \textbf{ 104.01 } & \textbf{ 77.91 } & \textbf{ 119.60 } & \textbf{ 77.42 } & \textbf{ 119.31 } & \textbf{ 69.74 } & \textbf{ 102.70 } \\
		\hline
	\end{tabular}}
	\label{h36m_long}
\end{table*}
\begin{table}[htbp]
	\centering
	\caption{Numbers of training parameters and efficiency.}
	\renewcommand{\arraystretch}{1}
	\setlength\tabcolsep{3.5pt}
	\resizebox{8.3cm}{0.88cm}
	{\begin{tabular}{cccc}
		\toprule
		Model & Traj-GCN & MSR-GCN & DMS-GCN \\
		\midrule
		num of parameters & 2.6M  & 6.3M  & \textbf{0.3M} \\
		time cost per epoch & 11.2min & 39.5min & \textbf{6.5min} \\
		\bottomrule
	\end{tabular}}
	\label{time}
\end{table}
\begin{figure}[!t]
	\centering
	\includegraphics[width=3.3in]{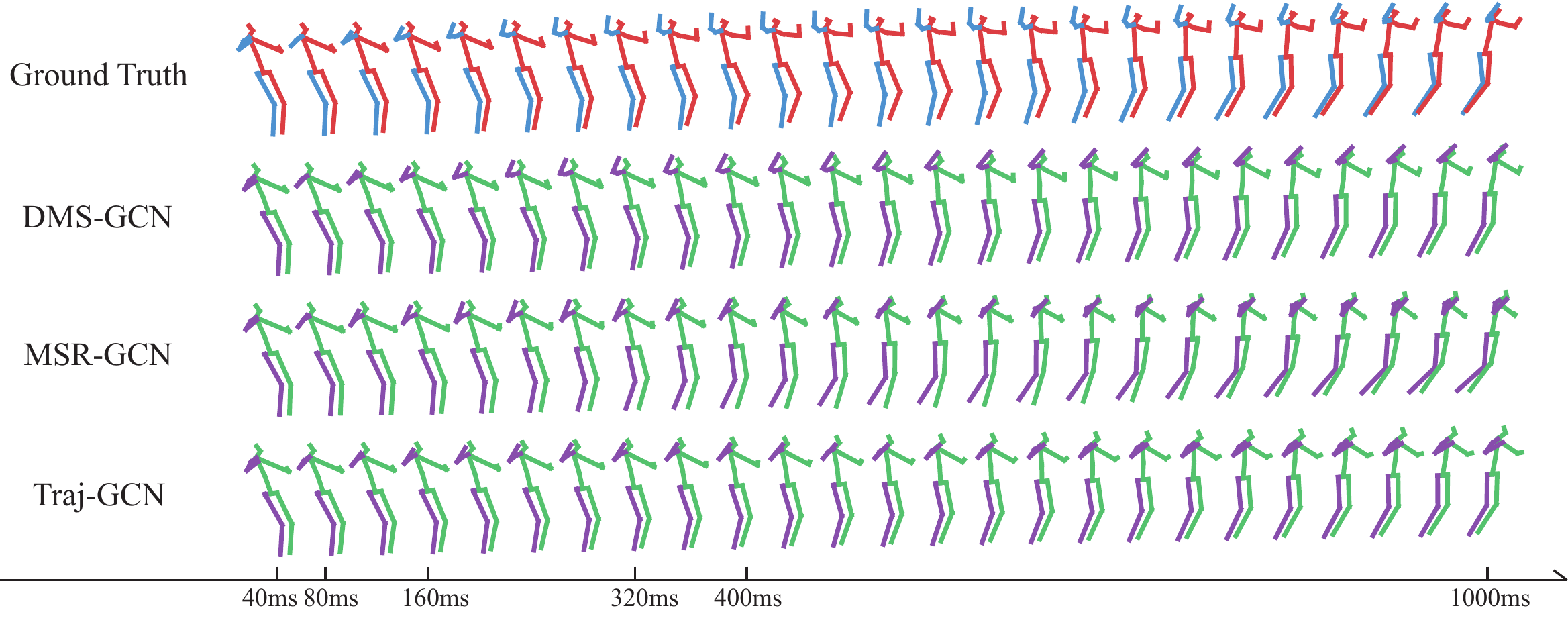}
	\caption{\textbf{Qualitative comparison.} We compare our method with two competitive ones by visualizing a sample of the H3.6M dataset. It is obvious that our approach (DMS-GCN) has the most convincing results.}
	\label{visualize}
\end{figure}
\begin{table}[htbp]
	\centering
	\caption{Comparisons for average error on CMU Mocap.}
	\renewcommand{\arraystretch}{1}
	\setlength\tabcolsep{3.5pt}
	\resizebox{8.3cm}{1.05cm}
	{\begin{tabular}{ccccccc}
			\toprule
			Millisecond (ms) & 80    & 160   & 320   & 400   & 560   & 1000 \\
			\midrule
			Residual sup. & 23.79  & 40.65  & 72.83  & 88.25  & 101.33  & 140.36  \\
			Traj-GCN & 10.17  & 18.54  & 34.62  & 42.11  & 55.32  & \textbf{ 81.29 } \\
			MSR-GCN & 9.23  & 16.05  & 31.28  & 39.03  & 56.17  & 84.84  \\
			DMS-GCN & \textbf{ 7.93 } & \textbf{ 14.45 } & \textbf{ 29.49 } & \textbf{ 37.90 } & \textbf{ 53.29 } & 82.13  \\
			\bottomrule
	\end{tabular}}
	\label{cmu}
\end{table}

\subsection{Loss and Evaluation Criteria}

\textbf{Loss function.} To train our model, we consider the $ \ell_{1} $ loss. Let the $ v $th joints at time $ t $ of the prediction sequence be $ \hat{\textbf{y}}_{t,v} $ and the corresponding ground truth be $ {\textbf{y}}_{t,v} $. The loss function is:
\begin{align}
	\mathcal{L}_{pred} = \frac{1}{VK}\sum_{t=T+1}^{T+K}\sum_{n=1}^{V}{\Vert \hat{\textbf{y}}_{t,v}-\textbf{y}_{t,v} \Vert}_{1}
\end{align}
where $ {\Vert \cdot \Vert}_{1} $ denotes the $ \ell_{1} $ norm. $ \ell_{1} $ loss gives significant gradients to joints with small losses and stable gradients to joints with large losses \cite{li2020dynamic}.
We found that $ \ell_{1} $ loss leads to more precise predictions than $ \ell_{2} $ loss does in our experiment.

\textbf{Evaluation Criteria.} Following the standard evaluation metric in \cite{mao2019learning,dang2021msr}, we report the results in Mean Per Joint Position Error (MPJPE) in milimeter, which is the most widely used evaluation metric:
\begin{align}
	\mathcal{L}_{MPJPE} = \frac{1}{VK}\sum_{t=T+1}^{T+K}\sum_{n=1}^{V}{\Vert \hat{\textbf{y}}_{t,v}-\textbf{y}_{t,v} \Vert}_{2}
\end{align}
where $ \hat{\textbf{y}}_{t,v} \in \mathbb{R}^{3} $ is the $ v $th joint position at time $ t $, and $ \textbf{y}_{t,v} $ is the corresponding ground truth. Instead of averaging the MPJPE loss along the time dimension, we also obtain the MPJPE loss at each time by calculating the MPJPE loss on every single frame.
\begin{table*}[htbp]
	\scriptsize
	\centering
	\caption{Ablation studies on the numbers of scales and GCNs vs. semi-learned GCNs. Results are the mean errors of 15 action categories on H3.6M. The best results are highlighted in bold. On average, all the designs of our model contribute to the accuracy.}
	\renewcommand{\arraystretch}{1}
	\setlength\tabcolsep{2pt}
	\resizebox{17.5cm}{1.2cm}
	{\begin{tabular}{l|c|cc|ccc|cccccc}
			\hline
			& TGCN  & SGCN   & SL-SGCN & Joint Scale & Bone Scale & Part Scale & 80    & 160   & 320   & 400   & 560   & 1000 \\
			\hline
			DMS-GCN-1L & \checkmark & & \checkmark & \checkmark &       &       & 11.08  & 23.98  & 50.68  & 62.68  & 82.56  & 118.18  \\
			DMS-GCN-2L & \checkmark &       & \checkmark & \checkmark & \checkmark &       & 10.68  & 23.97  & 51.75  & 63.88  & 83.27  & 117.89  \\
			DMS-GCN & \checkmark &       & \checkmark & \checkmark & \checkmark & \checkmark & \textbf{ 10.51 } & \textbf{ 23.61 } & \textbf{ 50.25 } & \textbf{ 62.35 } & \textbf{ 82.43 } & \textbf{ 116.93 } \\
			DMS-GCN w/o Mask & \checkmark & \checkmark &       & \checkmark & \checkmark & \checkmark & 11.08  & 24.06  & 51.17  & 63.07  & 82.54  & 117.13  \\
			DMS-GCN w/o TGCN &       &       & \checkmark & \checkmark & \checkmark & \checkmark & 13.75  & 29.14  & 59.27  & 71.91  & 90.88  & 123.28  \\
			\hline
	\end{tabular}}
	\label{ablation}
\end{table*}

\subsection{Results on Human3.6M Dataset}

Following previous work, we evaluate our model in 1000ms (i.e., 25 frames) high-precision long-term motion prediction. We compare our model with some competitive models \cite{martinez2017human,mao2019learning,dang2021msr} in qualitative and quantitive ways. It is worth mentioning that most works evaluate their methods on only eight sequences for each action category. We argue that such little test data is not enough to represent the performance of the compared approaches. \cite{dang2021msr} also questions on the same point. So we modified their published codes to use the whole test set.

\textbf{Qualitative results.} We first visualized the results of motion prediction sequences generated by different models (Traj-GCN \cite{mao2019learning}, MSR-GCN \cite{dang2021msr}, and ours) to make a qualitative comparison, as shown in Figure \ref{visualize}. Prediction sequences generated by other models show a visible difference compared with the ground truth. In this figure, the action we need to predict is to raise the right arm over the head and lower it a little. The MSR-GCN model does not foresee that the arm will be higher than the head, and the Traj-GCN outputs almost unmoved pose sequences. Though there still has a gap with the ground truth, our model (DMS-GCN) predicts the most reliable results.

\textbf{Quantitative results.} We also further evaluate the MPJPE error in 3D coordinates between the ground truth and some models (Residual sup\cite{martinez2017human}, Traj-GCN \cite{mao2019learning}, MSR-GCN \cite{dang2021msr}, and ours). The quantitative comparisons for both short-term and long-term prediction results are presented in Table \ref{h36m_short} and Table \ref{h36m_long} respectively. GCN-based approaches are better than the RNN-based method Residual sup. \cite{martinez2017human}, which validates the effectiveness of GCNs for human motion prediction. Among the methods based on GCNs, our method is superior to others, especially short-term prediction. As for the long-term, though the improvement is not obvious as the short-term, we still have the best results.

\textbf{Computational efficiency contrast.} The total number of training parameters as well as time cost per epoch are shown in Table \ref{time}. All experiments were performed using PyTorch on an NVIDIA Quadro RTX 6000 GPU. Our model (DMS-GCN) requires fewer parameters than existing methods, and its computation speed is significantly faster. We have the best prediction results while consuming the least computational resources compared to the two current best models, which shows the superiority of our model in all respects.

\subsection{Results on CMU Mocap Dataset}
Similar to the experiments above, we investigate the same method on the CMU Mocap dataset to prove the generalization capability of our model, as shown in Table \ref{cmu}. The experimental results show that our approach still achieves a significant improvement in the short-term prediction. To be honest, our model in the long-term prediction has not much advantage as usual and even exceeded by Traj-GCN slightly at the final frame. We argue two reasons: first, when predicting long-term poses, our model shares the same parameters with the short-term model, which causes the error accumulation; second, by visualizing the poses predicted by Traj-GCN, we see they tend to be a mean pose. Traj-GCN engages with minimizing the loss but neglects to capture the movement trend. In brief, our model still shows the most reliable results, which
proves that our model does not depend on specific actions or a specific number of human joints, i.e., our method is universal in the field of human motion prediction.

\subsection{Ablation Study}
\textbf{Effects of multi-scale architecture.} To study the effectiveness of the multi-scale mechanism, we conduct experiments on a different number of scales. The comparison results are shown in Table \ref{ablation}. Please see the rows corresponding to DMS-GCN-1L, DMS-GCN-2L, and DMS-GCN. The experiment with the whole three scales achieves a total victory. These experiments demonstrate the effectiveness of our multi-scale architecture.

\textbf{Effects of the semi-autonomous-learned GCNs.} To verify the effectiveness of the semi-autonomous-learned GCNs of the proposed architecture, we remove the mask for our model and find the results get worse at every sequence, as contrasted in the two rows corresponding to DMS-GCN and DMS-GCN w/o Mask of Table \ref{ablation}. This experiment illustrates the effectiveness of our semi-autonomous-learned GCNs.

\textbf{Effects of temporal GCN blocks.} We remove all TGCN blocks to analyze the effects of the TGCN blocks. Please see the rows corresponding to DMS-GCN and DMS-GCN w/o TGCN of Table \ref{ablation}. The experimental results show that TGCN blocks improve the prediction by a large margin, which strongly validates the importance of TGCN blocks.

\section{Conclusion and Future Work}

In this paper, we build dynamic multiscale spatiotemporal graph convolutional networks to effectively predict future human poses from observed histories. We also mask the adjacency matrix of spatial GCN blocks to make it semi-autonomous learned. We use a short observed pose sequence of 10 frames to predict 25 frames in the future. Our purposed method surpasses the two current best techniques while only requiring the fewest parameters. In our future work, we will consider the decoding capability of the decoder to perfect the long-term motion prediction.

\newpage

\bibliographystyle{named}
\bibliography{ijcai22}

\end{document}